\documentclass{article}
\usepackage[preprint,nonatbib]{neurips_2020}
\usepackage[T1]{fontenc}
\usepackage[utf8]{inputenc}
\usepackage{amsfonts}
\usepackage{amsmath,amsfonts,amssymb}
\usepackage{array}
\usepackage{bbm}
\usepackage{booktabs}
\usepackage{graphicx}
\usepackage{microtype}
\usepackage{nicefrac}
\usepackage{placeins}
\usepackage{subcaption}
\usepackage{url}
\usepackage{hyperref}
\usepackage[noabbrev]{cleveref}

\title{LSD-C\@: Linearly Separable Deep Clusters}

\author{Sylvestre-Alvise Rebuffi\thanks{indicates equal contribution} \hspace{2em} Sebastien Ehrhardt\footnotemark[1] \hspace{2em} Kai Han\footnotemark[1] \hspace{2em}\\
\textbf{Andrea Vedaldi \hspace{2em} Andrew Zisserman}\\
Visual Geometry Group, Department of Engineering Science, University of Oxford\\
\texttt{\{srebuffi,hyenal,khan,vedaldi,az\}@robots.ox.ac.uk} \\
}

\begin{document}
\maketitle
\begin{abstract}
We present LSD-C, a novel method to identify clusters in an unlabeled dataset.
Our algorithm first establishes pairwise connections in the feature space between the samples of the minibatch based on a similarity metric. Then it regroups in clusters the connected samples and enforces a linear separation between clusters. This is achieved by using the pairwise connections as targets together with a binary cross-entropy loss on the predictions that the associated pairs of samples belong to the same cluster.
This way, the feature representation of the network will evolve such that similar samples in this feature space will belong to the same linearly separated cluster.
Our method draws inspiration from recent semi-supervised learning practice and proposes to combine our clustering algorithm with self-supervised pretraining and strong data augmentation.
We show that our approach significantly outperforms competitors on popular public image benchmarks including CIFAR 10/100, STL 10 and MNIST, as well as the document classification dataset Reuters 10K. Our code is available at \url{https://github.com/srebuffi/lsd-clusters}.
\end{abstract}

\section{Introduction}\label{s:intro}

The need for large scale labelled datasets is a major obstacle to the applicability of deep learning to problems where labelled data cannot be easily obtained.
Methods such as clustering, which are unsupervised and thus do not require any kind of data annotation, are in principle more easily applicable to new problems.
Unfortunately, standard clustering algorithms~\cite{Comaniciu02meanshift, ester1996density, macqueen1967some, pearson1894contrib} usually do not operate effectively on raw data and require to design new data embeddings specifically for each new application.
Thus, there is a significant interest in automatically learning an optimal embedding while clustering the data, a problem sometimes referred to as simultaneous data clustering and representation learning.
Recent works have demonstrated this for challenging data such as images~\cite{ji2019invariant, xie2016unsupervised} and text~\cite{jiang2016variational, sarfraz2019efficient}.
However, most of these methods work with a constrained output space, which usually coincides with the space of discrete labels or classes being estimated, therefore forcing to work at the level of the semantic of the clusters directly.

In this paper, we relax this limitation by introducing a novel clustering method, \emph{Linearly Separable Deep Clustering} (LSD-C).
This method operates in the feature space computed by a deep network and builds on three ideas.
First, the method extracts mini-batches of input samples and establishes pairwise pseudo labels (connections) for each pair of sample in the mini-batch.
Differently from prior art, this is done in the space of features computed by the penultimate layer of the deep network instead of the final output layer, which maps data to discrete labels.
From these pairwise labels, the method learns to regroup the connected samples into clusters by using a clustering loss which forces the clusters to be linearly separable.
We empirically show in \cref{subsec:ablation} that this relaxation already significantly improves clustering performance.

Second, we initialize the model by means of a self-supervised representation learning technique.
Prior work has shown that these techniques can produce features with excellent linear linear separability~\cite{ chen2020simple, gidaris2018unsupervised, he2019momentum} that are particularly useful as initialization for downstream tasks such as semi-supervised and few-shot learning~\cite{gidaris2019boosting, rebuffi2019semi, zhai2019s4l}.

Third, we make use of very effective data combination techniques such as RICAP~\cite{takahashi2018ricap} and MixUp~\cite{zhang2017mixup} to produce composite data samples and corresponding pseudo labels, which are then used at the pairwise comparison stage.
In~\cref{sec:exp} we show that training with such composite samples and pseudo labels greatly improves the performance of our method, and is in fact the key to good performance in some cases.

We comprehensively evaluate our method on popular image benchmarks including CIFAR 10/100, STL 10 and MNIST, as well as the document classification dataset Reuters 10K.
Our method almost always outperforms competitors on all datasets, establishing new state-of-the-art clustering results.
The rest of the paper is organized as follows.
We first review the most relevant works in section~\ref{sec:related}. Next, we develop the details of our proposed method in section~\ref{sec:method}, followed by the experimental results, ablation studies and analysis in section~\ref{sec:exp}. Our code is publicly available at \url{https://github.com/srebuffi/lsd-clusters}.
\section{Related work}
\label{sec:related}
\textbf{Deep clustering}. Clustering has been a long-standing problem in the machine learning community, including well-known algorithms such as K-means~\cite{macqueen1967some}, mean-shift~\cite{Comaniciu02meanshift}, DBSCAN~\cite{ester1996density} or Gaussian Mixture models~\cite{pearson1894contrib}. Furthermore it can also be combined with other techniques to achieve very diverse tasks like novel category discovery~\cite{han19DTC,fontanel2020boosting} or semantic instance segmentation~\cite{de2017semantic} among others. With the advances of deep learning, more and more learning-based methods have been introduced in the literature~\cite{genevay2019differentiable,ghasedi2017deep,haeusser2018associative, huang2019centroid, jiang2016variational,li2018discriminatively,shaham2018spectralnet,xie2016unsupervised,yang2017towards}). Among them, DEC \cite{xie2016unsupervised} is one of the most promising method. It is a two stage method that jointly learns the feature embedding and cluster assignment. The model is first pretrained with an autoencoder using reconstruction loss, after which the model is trained by constructing a sharpened version of the soft cluster assignment as pseudo target. This method inspired a few following works such as IDEC~\cite{guo2017improved} and
DCED~\cite{guo2017deep}.
JULE \cite{yang2016joint} is a recurrent deep clustering framework that jointly learns the feature representation with an agglomerative clustering procedure, however it requires tuning a number of hyper-parameters, limiting its practical use. More recently, several methods have been proposed based on mutual information~\cite{chen2016infogan,hu2017learning,ji2019invariant}. Among them, IIC~\cite{ji2019invariant} achieves the current state-of-the-art results on image clustering by maximizing the mutual information between two transformed counterparts of the same image.
Closer to our work is the DAC~\cite{chang2017deep} method, which considers clustering as a binary classification problem. By measuring the cosine similarity between predictions, pairwise pseudo labels are generated from the most confident positive or negative pairs. With the generated pairwise pseudo labels, the model can then be trained by a binary cross-entropy loss. DAC can learn the feature embedding as well as the cluster assignment in an end-to-end manner. Our work significantly differs from DAC as it generates pairwise predictions from a less constrained feature space using similarity techniques not limited to cosine distance.

\textbf{Self-supervised representation learning}. Self-supervised representation learning has recently attracted a lot of attention. Many effective self-supervised learning methods have been proposed in the literature~\cite{asano2019self, caron2018deep,chen2020simple, gidaris2020learning,gidaris2018unsupervised,he2019momentum}. DeepCluster \cite{caron2018deep} learns feature representation by classification using the pseudo labels generated from K-means on the learned features in each training epoch. RotNet~\cite{gidaris2018unsupervised} randomly rotates an image, and learns to predict the applied rotations. Very recently, contrastive learning based methods MoCo~\cite{he2019momentum} and SimCLR \cite{chen2020simple} have achieved the state-of-the-art self-supervised representation performance, surpassing the representation learnt using ImageNet labels. Self-supervised learning has been also applied in few-shot learning~\cite{gidaris2019boosting}, semi-supervised learning~\cite{rebuffi2019semi,zhai2019s4l} and novel category discovery~\cite{han2020automatically}, which successfully boosts their performance. In this work we make use of the provably well-conditioned feature space learnt from self-supervised learning method to initialize our network and avoid degenerative cases.

\textbf{Pairwise pseudo labeling}. Pairwise similarity between pairs of sample has been widely used in the literature for dimension reduction or clustering (e.g., t-SNE~\cite{maaten2008visualizing}, FINCH~\cite{sarfraz2019efficient}). Several methods have shown the effectiveness of using pairwise similarity to provide pseudo labels on-the-fly to train deep convolutional neural networks. In~\cite{hsu2019multi}, a binary classifier is trained to provide pairwise pseudo labels to train a multi-class classifier. In~\cite{han2020automatically}, ranking statistics is used to obtain pairwise pseudo labels on-the-fly for the task of novel category discovery. In~\cite{sarfraz2019efficient}, the pairwise connection between data points by finding the nearest neighbour is used to cluster images using CNN features. In our method, we compute pairwise labels from a neural network embedding. This way we generate pseudo labels for each pair in each mini-batch and learn cluster assignment without any supervision.

\section{Method}\label{sec:method}

\begin{figure}
  \centering
  \includegraphics[width=\linewidth]{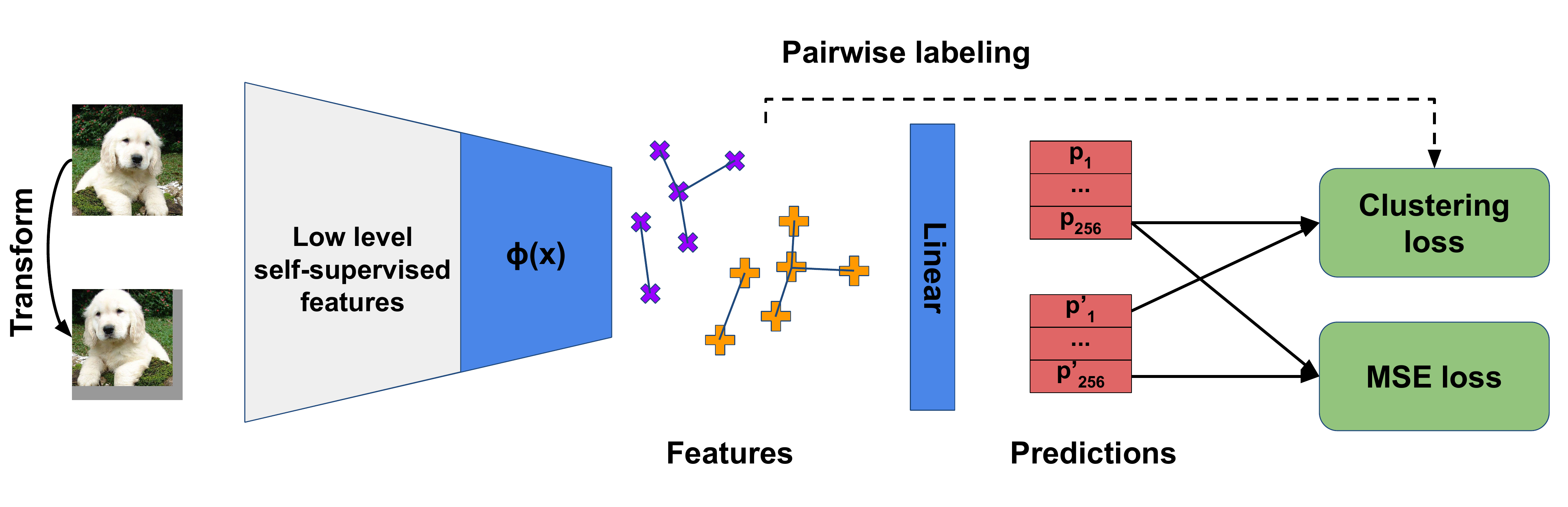}
  \caption{\textbf{Overview of LSD-C.} Pairwise labels are extracted at the feature level. They are then used in a clustering loss after the linear classifier. This way, the feature maps will evolve such that connected samples will be grouped in linearly separated clusters. The MSE loss acts a regularizer and enforces the consistency of the cluster predictions when data augmentation is applied.}
  \label{fig:overview}
\end{figure}

Our methods is divided into three stages: (i) self-supervised pre-training, (ii) pairwise connection and clustering, and (iii) data composition.
We provide an overview of our pipeline in \cref{fig:overview}.
Our method processes each input data batch $x$ in two steps, by extracting features $\mathbf{f} =\Phi(x) \in \mathbb{R}^{N\times D}$ by means of a neural network $\Phi$, followed by estimating posterior class probabilities $\mathbf{p} = \Psi(\mathbf{f})\in \mathbb{R}^{N\times K}$ by means of a linear layer $\Psi$ and softmax non-linearity.
We use the symbol $\mathbf{p}' = \Phi(\Psi(x'))$ to denote the class predictions for the same mini-batch $x'$ with data augmentation (random transformations) applied to it.
We use the letters $D$, $K$ and $N$ to denote the feature space dimension, the number of clusters and the mini-batch size.
We now detail each component of LSD-C.


\subsection{Self-supervised pretraining}

As noted in the introduction, traditional clustering methods require handcrafted or pretrained features.
More recently, methods such as~\cite{ji2019invariant} have combined deep learning and clustering to learn features and clusters together;
even so, these methods usually \emph{still} require ad hoc pre-processing steps (e.g.~pre-processing such as Sobel filtering~\cite{caron2018deep, ji2019invariant}) and extensive hyperparameter tuning.
In our method we address this issue and avoid bad local minima in our clustering results by initializing our representation by means of self-supervised learning.
In practice, this amounts to train our model on a pretext task (detailed in \cref{sec:exp}) and then retain and freeze the earlier layers of the model when applying our clustering algorithm.
As reported in~\cite{chen2020simple, gidaris2018unsupervised}, the features obtained from self-supervised pre-training are linearly separable with respect to typical semantic image classes.
This property is particularly desirable in our context and also motivates our major design choice: since the feature space of self-supervised pre-trained network is linearly separable, it is therefore easier to directly operate on it to discriminate between different clusters.

\subsection{Pairwise labeling}

A key idea in our method is the choice of space where pairwise the data connections are established:
we extract pairwise labels at the level of the data representation rather than at the level of the class predictions.
The latter is a common design choice, used in DAC~\cite{chang2017deep} to establish pairwise connections between data points and in DEC~\cite{xie2016unsupervised} to match the current label posterior distribution to a sharper version of itself.

The collection of pairwise labels between samples in a mini-batch is given by the adjacency matrix $A$ of an undirected graph whose nodes are the samples and whose edges encode their similarities.
DAC~\cite{chang2017deep} generates pseudo labels by checking if the output of the network is above or under certain thresholds.
The method of~\cite{lee2013pseudo} proceeds similarly in the semi-supervised setting.
In our method, as we work instead at the feature space level, the pairwise labeling step is a separate process from class prediction and we are free to choose any similarity to establish our adjacency matrix $A$.
We denote with $\mathbf{f}_i \in \mathbb{R}^D$ and $\mathbf{f}_j\in \mathbb{R}^D$ the feature vectors for samples $i$ and $j$ in a mini-batch, obtained from the penultimate layer of the neural network $\Phi$.
We also use the symbol $A_{ij} \in \{0,1\}$ to denote the value of the adjacency matrix for the pair of samples $(i,j)$. Next, we describe the different types of pairwise connections considered in this work and summarize them in \cref{table:pairwise_lab}.

\paragraph{Cosine  and $L_2$ similarity.}

Let $\tau \in \mathbb{R}^+$ be a threshold hyperparameter and define
$
C_{ij} = [\cos(\mathbf{f}_j,\mathbf{f}_i) > \tau]
$
(cosine)
or
$
C_{ij} = [\| \mathbf{f}_j - \mathbf{f}_i \|^2 < \tau]
$
(Euclidean)
where $\cos$ denotes the dot product between $L_2$-normalized vectors.
We then define
$
A_{ij} =  \mathbbm{1}_{C_{ij}}
$
where $\mathbbm{1}$ is the indicator function.
These definitions connect neighbor samples but do not account well for the local structure of the data.
Indeed, it is not obvious that the cosine similarity or Euclidean distance would establish good data connections in feature space.

\paragraph{Symmetric SNE.}

\begin{figure}
\centering
\begin{subfigure}{.24\textwidth}
  \centering
  \includegraphics[width=1.\linewidth]{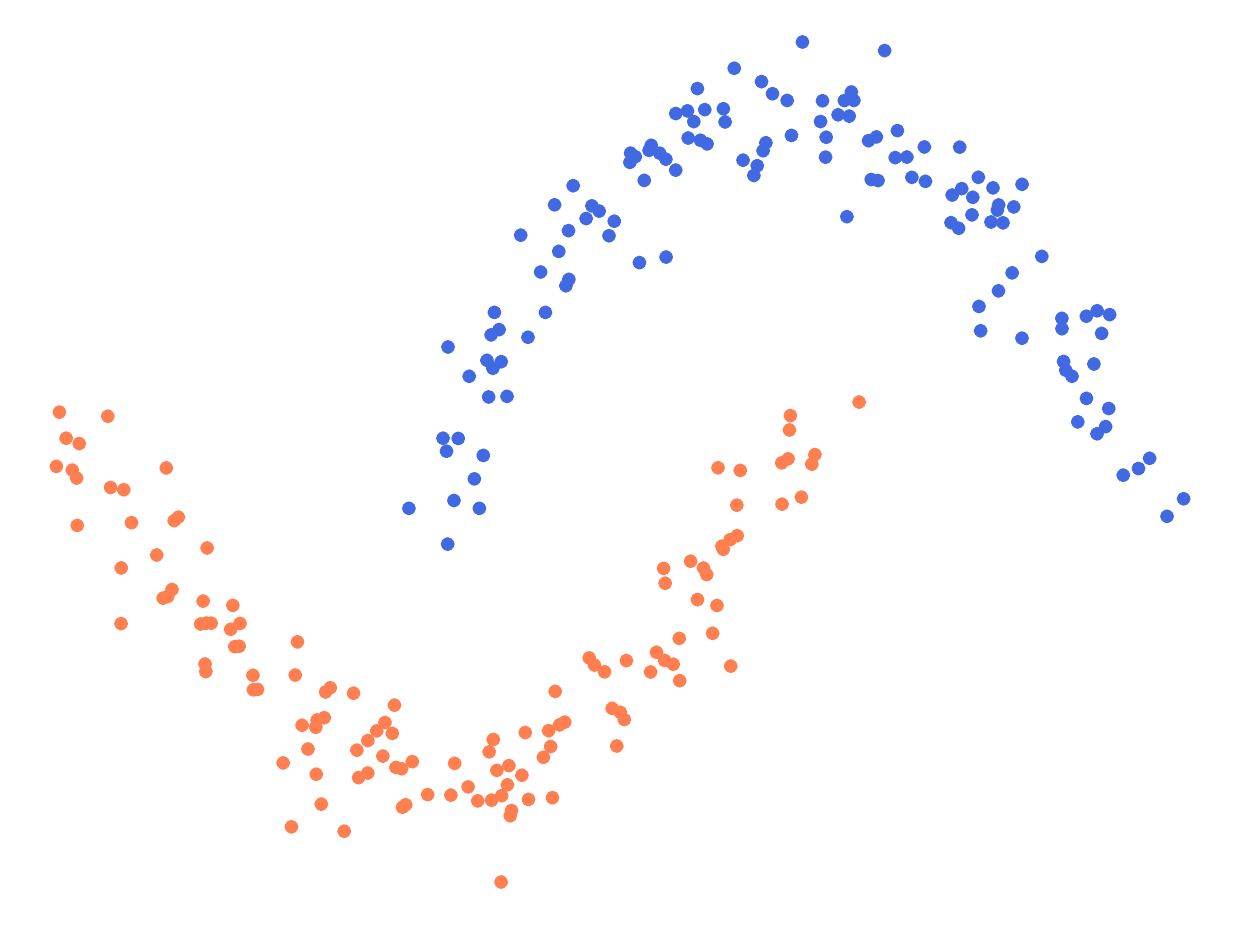}
  \caption{Raw data}
  \label{fig:raw_two_moons}
\end{subfigure}
\begin{subfigure}{.24\textwidth}
  \centering
  \includegraphics[width=1.\linewidth]{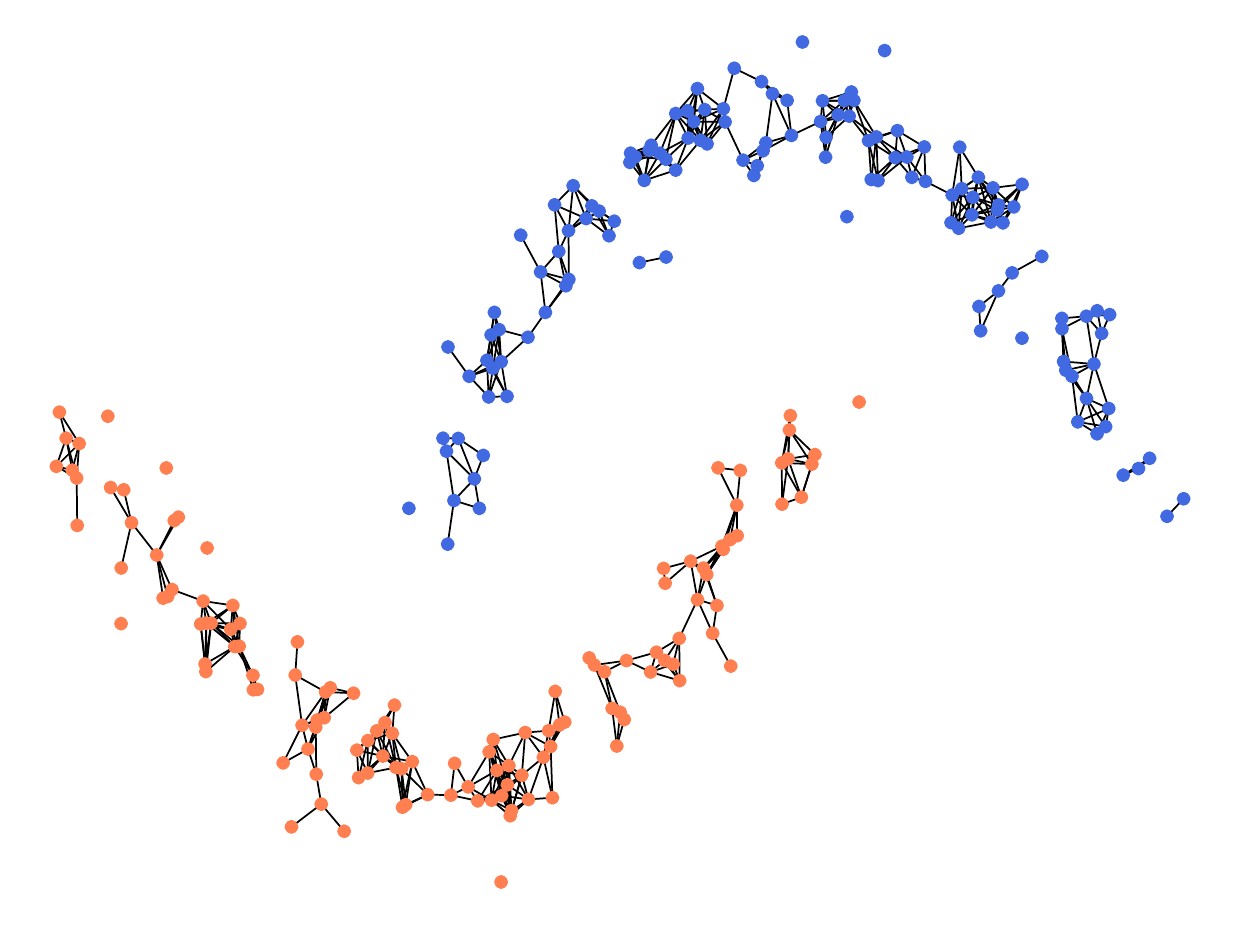}
  \caption{$L_2$ dist.}
  \label{fig:labeling_l2}
\end{subfigure}%
\begin{subfigure}{.24\textwidth}
  \centering
  \includegraphics[width=1.\linewidth]{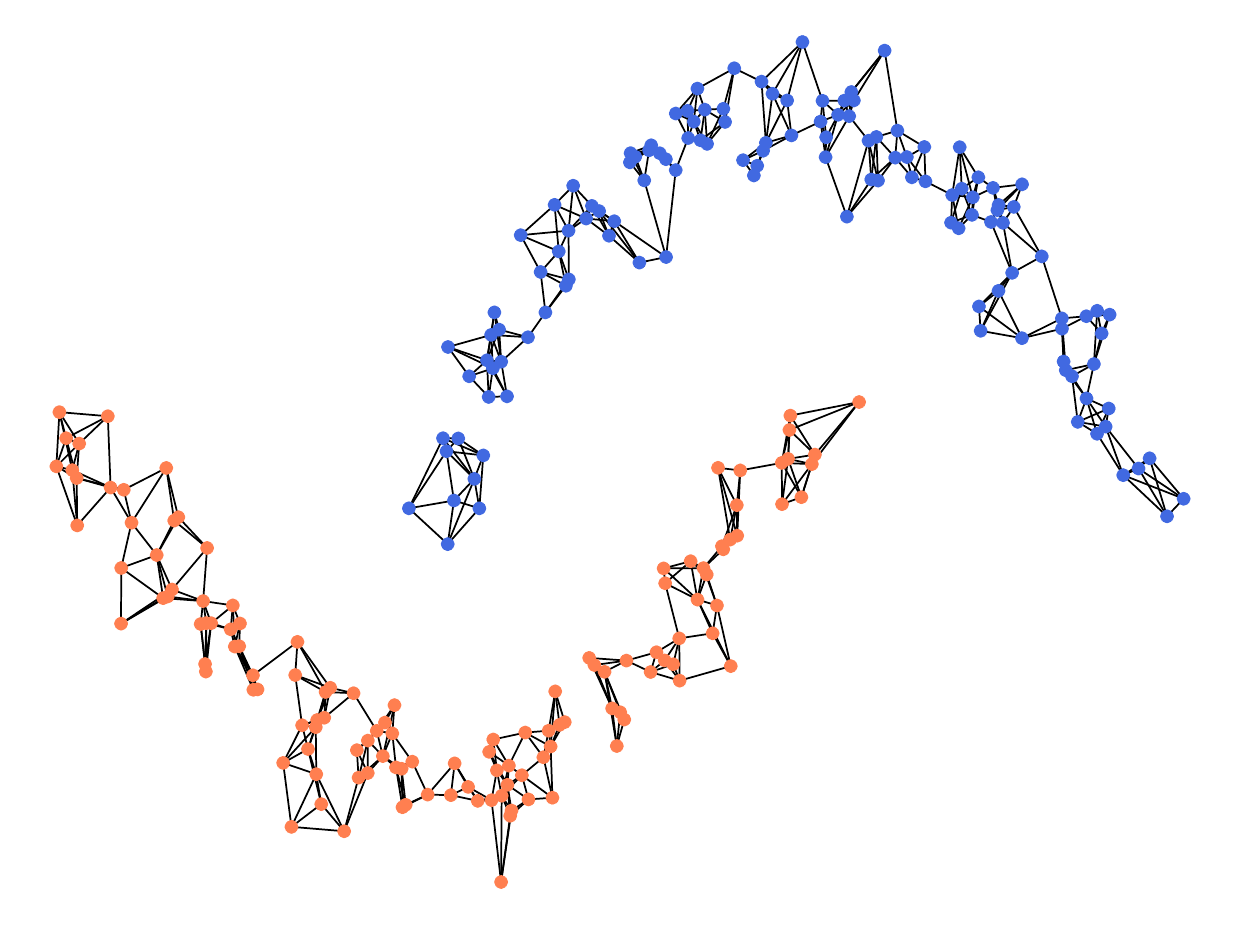}
  \caption{kNN}
  \label{fig:labeling_knn}
\end{subfigure}
\begin{subfigure}{.24\textwidth}
  \centering
  \includegraphics[width=1.\linewidth]{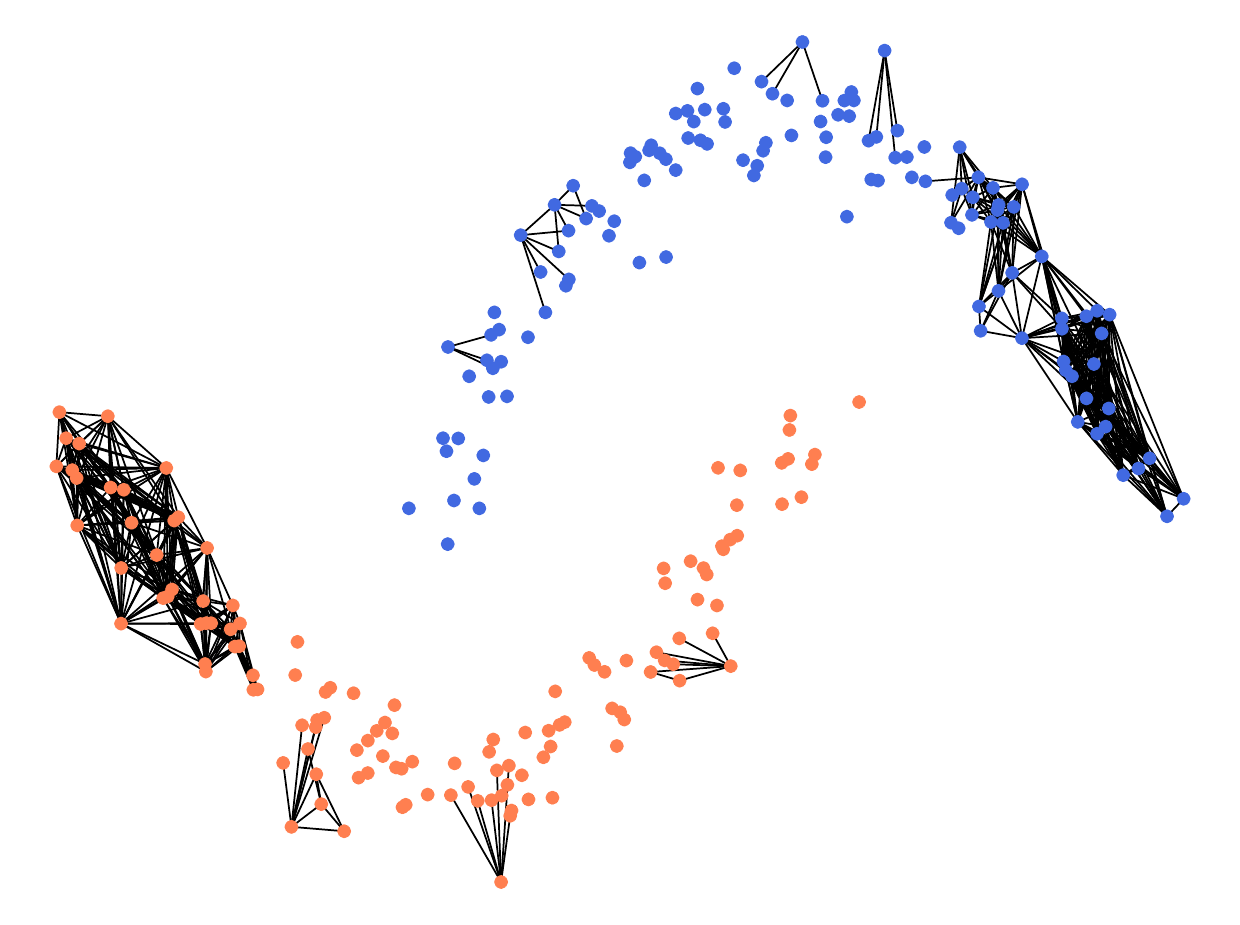}
  \caption{SNE}
  \label{fig:labeling_tsne}
\end{subfigure}
\caption{ \textbf{Pairwise connections on the two moons toy data.} From left to right. We apply our algorithm with different connection techniques on a toy dataset shown in (a) where each color represents a class. We use the different connections techniques of \cref{table:pairwise_lab} such that there are 650 undirected edges for each similarity. Compared to $L_2$ distance and SNE, kNN produces neighbourhoods of similar sizes and every sample is connected. SNE captures the local structure of the data: most of the connections are at the external tails of the moons where there are less points.}
\label{fig:labeling}
\end{figure}

A possible solution to alleviate the previous issue is to use the symmetric SNE similarity introduced in t-SNE~\cite{maaten2008visualizing}.
This similarity is based on the conditional probability $p_{j\mid i}$ of picking $j$ as neighbor of $i$ under a Gaussian distribution assumption.
We make a further assumption compared to~\cite{maaten2008visualizing} of an equal variance for every sample in order to speed up the computation of pairwise similarities and define:
\begin{equation}
p_{j\mid i} = \frac{\exp(-\| \mathbf{f}_j - \mathbf{f}_i \|^2/T^2)}{\sum\limits_{k \neq i} \exp(-\| \mathbf{f}_k - \mathbf{f}_i \|^2/T^2)} = \frac{\exp(-\| \mathbf{f}_j - \mathbf{f}_i \|^2/T^2)}{Z_i},
\label{eq:cond_prob}
\end{equation}
\vspace{-0.7em}
\begin{equation}
C_{ij} = \frac{p_{j|i} + p_{i|j}}{2} > \tau
~~~\Longleftrightarrow~~~
\frac{\exp(-\| \mathbf{f}_j - \mathbf{f}_i \|^2/T^2)}{H(Z_i, Z_j)} > \tau.
\label{eq:adj_tsne}
\end{equation}
As shown in equation~\eqref{eq:cond_prob}, we introduce a temperature hyperparameter $T \in \mathcal{R}^+$ and we call $Z_i$ the partition function for sample $i$. Then the associated adjacency matrix in equation~\eqref{eq:adj_tsne} can be written as a function of the $L_2$ distance between samples and, in the denominator, of the harmonic mean $H$ of the partition functions.
 As a result, if sample $i$ or $j$ has many close neighbours, it will reduce the symmetric SNE similarity and possibly prevent a connection between samples $i$ and $j$.
 Such a phenomenon is shown on the two moons toy dataset in \cref{fig:labeling}.

\paragraph{k-nearest neighbors.} We also propose a similarity based on $k$-nearest neighbours (kNN)~\cite{cover1967nearest} where the samples $i$ and $j$ are connected if $i$ is in the $k$-nearest neighbours of $j$ or if $j$ is in the $k$-nearest neighbours of $i$. With this similarity, the hyperparameter is the minimum of neighbours $k$ and not the threshold $\tau$.

\begin{table}[t]
  \caption{\textbf{Pairwise labeling with  adjacency matrices $A_{ij} =  \mathbbm{1}_{C_{ij}}$ based on different similarities.} $\tau$ is the thresholding hyperparameter for $L_2$, SNE and Cosine. The number of neighbours $k$ is kNN's hyperparameter.}
  \label{table:pairwise_lab}
  \centering
  \begin{tabular}[c]{p{0.4cm}cccc}
    \toprule
                 & $L_2$ dist. & SNE & Cosine & kNN \\
    \midrule
    $C_{ij} = $     & ~   $ \| \mathbf{f}_j - \mathbf{f}_i \|^2 < \tau$ & $\frac{\exp(-\| \mathbf{f}_j - \mathbf{f}_i \|^2/T^2)}{H(Z_i, Z_j)} > \tau$ & $ \frac{\mathbf{f}_j^\top \mathbf{f}_i}{\| \mathbf{f}_j \| \| \mathbf{f}_i \|} > \tau$ &  $(j \in \text{kNN}(i)) \lor (i \in \text{kNN}(j))$ \\
    \bottomrule
  \end{tabular}
\end{table}

\subsection{Clustering loss and data composition}

Now that we have established pairwise connections between each pair of samples in the mini-batch, we will use the adjacency matrix as target for a binary cross-entropy loss.
Denoting with $P(i=j)$ the probability that samples $i$ and $j$ belong to the same cluster, we wish to optimize the clustering loss:
\begin{equation}\label{eq_loss}
  L_\text{clus} =
  -\sum\limits_{i, j}
  A_{ij} \log P(i=j)
  +
  (1 -  A_{ij}) \log P(i \neq j).
\end{equation}
The left term of this loss aims at maximizing the number of connected samples (i.e.~$A_{ij} = 1$) within a cluster and the right term at minimizing the number of non-connected samples within it (namely, the edges of the complement of the similarity graph $1 - A_{ij} = 1$).
Hence the second term prevents the formation of a single, large cluster that would contain all samples.


The next step is to model $P(i=j)$ by using the linear classifier predictions of samples $i$ and $j$.
As seen in equation \eqref{eq_proba}, for a fixed number of clusters $K$, the probability of samples $i$ and $j$ belonging to the same cluster can be rewritten as a sum  of probabilities over the possible clusters.
For simplicity, we assume that samples $i$ and $j$ are independent.
This way, the pairwise comparison between samples appear only at the loss level and we can thus use the standard forward and backward passes of deep neural networks where each sample is treated independently.
By plugging equation~\eqref{eq_proba} in equation~\eqref{eq_loss} and by replacing $\mathbf{p}_j$ with $\mathbf{p}'_j$ to form pairwise comparisons between the mini-batch and its augmented version, we obtain our final clustering loss $\mathcal{L}_\text{clus}$:
\begin{equation}
  P(i=j) = \sum\limits_{k=1}^K P(i=k, j=k) = \sum\limits_{k=1}^K P(i=k) P(j=k) = \mathbf{p}_i^\top \mathbf{p}_j,
\label{eq_proba}
\end{equation}
\vspace{-0.5em}
\begin{equation}
L_\text{clus} = -\sum\limits_{i, j} A_{ij} \log(\mathbf{p}_i^\top \mathbf{p}'_j) + (1 -  A_{ij}) \log(1 - \mathbf{p}_i^\top \mathbf{p}'_j).
\label{eq_loss_2}
\end{equation}
A similar loss is used in~\cite{hsu2019multi} but with supervised pairwise labels to transfer a multi-class classifier across tasks.
It is also reminiscent of DAC~\cite{chang2017deep}, but differs from the latter because the DAC loss does not contain a dot product between probability vectors but between $L_2$ normalized probability vectors.
Hence DAC optimizes a Bhattacharyya distance whereas we optimize a standard binary cross-entropy loss.

\begin{figure}
  \centering
  \includegraphics[width=\linewidth]{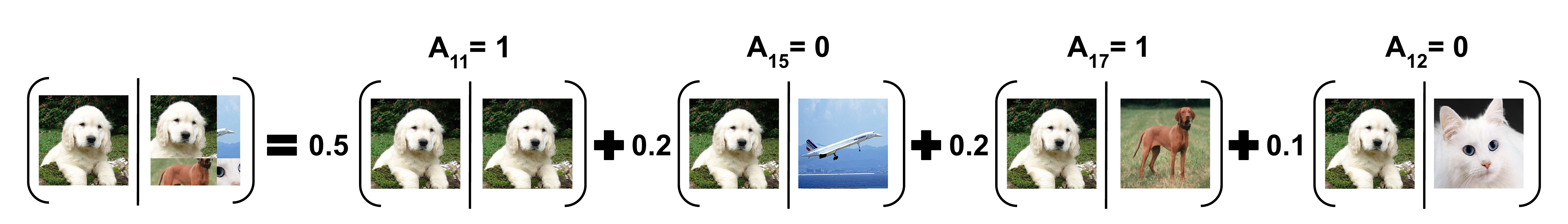}
  \caption{Illustration for~\cref{eq_loss_aug} of a pairwise target between the "pure" image $i=1$ and the composite image $j$ with $\sigma(j) \in (1, 5, 7, 2)$. In this case, the resulting pairwise target equals $0.7$.}
  \label{fig:ricap}
\end{figure}

In practice $L_\text{clus}$ can be used in combination with effective data augmentation techniques such as RICAP~\cite{takahashi2018ricap} and MixUp~\cite{zhang2017mixup}.
These methods combine the images from the minibatch and use a weighted combination of the labels of the original images as new target for the cross-entropy loss.
We denote with $\sigma$ permutation of the samples in the minibatch; RICAP and MixUp require 4 and 2 permutations respectively.
RICAP creates a new minibatch of composite images by patching together random crops from the 4 permutations of the original minibatch, whereas MixUp produces a new minibatch by taking a linear combination with random weights from 2 permutations.
The new target for a composite image is then obtained by taking a linear combination of the labels in the recombined images, weighted by area proportions in RICAP and the mixing weights in MixUp.
These techniques were proposed for the standard supervised classification setting, so we adapt them here to clustering.
In order to do so, we propose to perform a pairwise labeling between the composite images and the raw original images.
Both minibatches of original and composite images are fed to the network.
Then, as illustrated in \cref{fig:ricap}, the pairwise label between a composite image and a raw image is the linear combination of the pairwise labels between the components of both.
To sum up, to obtain the pairwise labels between a minibatch and its composite version we just need to extract the adjacency matrix $A$ of the minibatch and then do a linear combination of the adjacency matrix $A$ with the different column permutations $\sigma$:
\begin{equation}
  L_\text{clus} = -\sum\limits_{\sigma}\sum\limits_{i, j} w_\sigma A_{i\sigma(j)} \log(\mathbf{p}_i^\top \Tilde{\mathbf{p}}_j) + (1 -  w_\sigma A_{i\sigma(j)}) \log(1 - \mathbf{p}_i^\top \Tilde{\mathbf{p}}_j)
  \label{eq_loss_aug}
  \end{equation}
Regarding the predicted probability of the `pure' image $i$ and the composite image $j$ being in the same cluster, we take the dot product between their respective cluster predictions $\mathbf{p}_i$ and $\Tilde{\mathbf{p}}_j$.

\subsection{Overall loss}

The overall loss we optimise is given by
\begin{equation}
L_\text{tot} = L_\text{clus}(\mathbf{f}, \mathbf{p}, \mathbf{p}') + L_\text{cons}(\mathbf{p}, \mathbf{p}'),
\label{eq:tot_loss}
\end{equation}
where
\begin{equation}
L_\text{cons} = \frac{\omega(t)}{KN} \sum_{i=1}^{N} \| \mathbf{p}_i -  \mathbf{p}'_i \|^2,
\label{eq:cons}
\end{equation}
and $\omega(t) = \lambda e^{-5(1-\frac{t}{T})^{2}}$ is the  ramp-up function proposed in~\cite{laine2016temporal, tarvainen2017mean} with $t$ the current training step, $T$ the ramp-up length and $\lambda \in \mathbb{R}_+$.
$L_\text{cons}$ is a consistency constraint which requires the model to produce the same prediction $\mathbf{p}\approx\mathbf{p}'$ for an image and an its augmented version.
We use it in our method in a similar way as semi-supervised learning techniques~\cite{laine2016temporal, miyato2018virtual, sajjadi2016regularization, tarvainen2017mean}, i.e.~as a regularizer to provide consistent predictions.
This differs significantly from clustering methods like IIC~\cite{ji2019invariant} and IMSAT~\cite{hu2017learning} where augmentations are used as a main clustering cue by maximizing the mutual information between different versions of an image.
Instead, as commonly done in semi-supervised learning, we use the Mean Squared Error (MSE) between predictions as the consistency loss.

\section{Experiments}
\label{sec:exp}
\paragraph{Datasets.}

We conduct experiments on five popular benchmarks which we use to compare our method against recent state-of-the-art approaches whenever results are available.
We use four image datasets and one text dataset to illustrate the versatility of our approach to different types of data. We use MNIST~\cite{lecun1998gradient}, CIFAR 10~\cite{Krizhevsky09cifar}, CIFAR 100-20~\cite{Krizhevsky09cifar} and STL 10~\cite{coates2011analysis} as image datasets.
All these datasets cover a wide range of image varieties ranging from $28\times 28$ pixels grey scale digits in MNIST to $96 \times 96$ higher resolution images from STL 10.
CIFAR 100-20 is redesigned from original CIFAR 100 since we consider only the 20 meta classes for evaluation as common practice~\cite{ji2019invariant}.
Finally we also evaluate our method on a text dataset, Reuters 10K~\cite{lewis2004rcv1}.
Reuters 10K contains 10,000 English news labelled with 4 classes.
Each news has 2,000 \emph{tf-idf} features.
For all datasets we suppose the number of classes to be known.

\paragraph{Experimental details.}

We use ResNet-18~\cite{he2016deep} for all the datasets except two. For MNIST we use a model inspired from VGG-4~\cite{simonyan2014very}, described in~\cite{ji2019invariant} and for Reuters 10K we consider a simple DNN of dimension 2000–500–500–2000–4 described in~\cite{xie2016unsupervised}.
We train with batch-size of 256 for all experiments. We use SGD optimizer with momentum~\cite{sutskever2013importance} and weight decay set to $5 \times 10^{-4}$ for every dataset except for Reuters 10K where we respectively use Adam~\cite{kingma2014adam} and decay of $2 \times 10^{-3}$. When comparing with other methods in \cref{tab:main-res} and~\cref{tab:reut}, we run our method using 10 different seeds and report average and standard deviation on each dataset to measure the robustness of our method with respect to initialization.
As it is common practice~\cite{ji2019invariant}, we train and test the methods on the whole dataset (this is acceptable given that the method uses no supervision).
Further experimental details about data augmentation and training are available in the appendix.

\paragraph{Evaluation metrics.}

We take the commonly used \emph{clustering accuracy} (ACC) as evaluation metric.
ACC is defined as
\begin{equation}
\max _{g \in \operatorname{Sym}\left(K\right)}
\frac{1}{N} \sum_{i=1}^{N}
\mathbbm{1}\left \{
  \overline{y}_{i}=g\left(y_{i}\right)
\right \},
\end{equation}
where $\overline{y}_{i}$ and $y_i$ respectively denote the ground-truth class label and the clustering assignment obtained by our method for each sample in the dataset. $\operatorname{Sym}\left(K\right)$ is the group of permutations with $K$ elements and following other clustering methods we use the Hungarian algorithm~\cite{kuhn1955hungarian} to optimize the choice of permutation.

\subsection{Results on standard benchmarks}
\begin{table}[t]
  \caption{{\bf Comparison with other methods.} Our method almost constantly reaches state-of-the-art performances by a large margin. Note that \cite{ji2019invariant} report best results over all the heads while we report results over ten different initializations. This further shows that our method is overall stable and robust to initialization.}
  \label{tab:main-res}
  \centering
 \footnotesize
  \begin{tabular}[c]{lcccccc}
    \toprule
                 & K-means~\cite{macqueen1967some}  & JULE~\cite{yang2016joint} & DEC~\cite{xie2016unsupervised} & DAC~\cite{chang2017deep} & IIC~\cite{ji2019invariant} & Ours\\
    \midrule
    CIFAR 10     &   22.9   & 27.2 & 30.1 &  52.2 & 61.7 & \textbf{81.7} $\pm$ 0.9  \\
    CIFAR 100-20 &   13.0   & 13.7 & 18.5 & 23.8  & 25.7 & \textbf{42.3} $\pm$ 1.0 \\
    STL 10       &   19.2   & 27.7 & 35.9 & 47.0  & 59.6 & \textbf{66.4} $\pm$ 3.2 \\
    MNIST        &   57.2   & 96.4 & 84.3 & 97.8  & \textbf{99.2} &  98.6 $\pm$ 0.5  \\
    \bottomrule
  \end{tabular}
\end{table}

We compare our method with the K-means~\cite{macqueen1967some} baseline and recent clustering methods. In \cref{tab:main-res}, we report results on image datasets.
We use RotNet~\cite{gidaris2018unsupervised} self-supervised pre-training for each dataset on all the data available (\emph{e.g} including the unlabelled set in STL-10).
Our method significantly outperforms the others by a large margin.
For example, our method achieves $81.5\%$ on CIFAR 10, while the previous state-of-the-art method IIC~\cite{ji2019invariant} gives $61.7\%$.
On CIFAR 10, our method also outperforms the leading semi-supervised learning technique FixMatch~\cite{sohn2020fixmatch} which obtains $64.3\%$ in its one label per class setting.
Similarly, on CIFAR 100-20 and STL 10, our method outperforms other clustering approaches respectively by $14.7$ and $6.8$ points.
On MNIST, our method and IIC both achieve a very low error rate around $1\%$.

These results clearly show the effectiveness of our approach.
Unlike the previous state-of-the-art method IIC that requires to apply Sobel filtering and very large batch size during training, our method does not require such preprocessing and works with a common batch size.
We also note that our method is robust to different initialization, with a maximum $3.2\%$ of standard deviation across all datasets.

To analyse further the results on CIFAR 10, we can look at the confusion matrix resulting from our model's predictions.We note that most of the errors are due to the `cat' and `dog' classes being confused.
If we retain only the confident samples with prediction above $0.9$ (around $60 \%$ of the samples), the accuracy rises to $94\%$.
We assume that the two classes `cat' and `dog' are are more difficult to discriminate due to their visual similarity.

In \cref{tab:reut}, we also evaluate our method on the document classification dataset Reuters 10K to show its versatility.
We compare with different approaches than in \cref{tab:main-res} as clustering methods developed for text are seldom evaluated on image datasets like CIFAR and vice versa.
Following existing approaches applied to Reuters 10K, we pretrain the deep neural network by training a denoising autoencoder on the dataset~\cite{jiang2016variational}.
Our method works notably better than the K-means baseline, and is on par with the best results methods FINCH~\cite{sarfraz2019efficient} and VaDE~\cite{jiang2016variational}.
Most notably one run of our method established state-of-the-art results of 83.5\%, 2 points above the current best model.

\begin{table}[t]
  \caption{\textbf{Results on Reuters 10K.} Our method performs on average on par with state of the art. Note that for the best seed we reach state-of-the-art results of $83.5\%$.}
  \label{tab:reut}
  \centering
 \footnotesize
  \begin{tabular}[c]{lcccccc}
    \toprule
                 & 
     K-means~\cite{macqueen1967some}  &
     IMSAT~\cite{hu2017learning} & DEC~\cite{xie2016unsupervised} &  VaDE~\cite{jiang2016variational} &
     FINCH~\cite{sarfraz2019efficient} &
     Ours\\
    \midrule
    Reuters 10K     &  52.4  & 71.9   & 72.2 & 79.8 & \textbf{81.5} & 79.0  $\pm$ 4.3  \\
    \bottomrule
  \end{tabular}
\end{table}

\subsection{Ablation studies}\label{subsec:ablation}

In order to analyze the effects of the different components of our method, we conduct a three parts ablation study on CIFAR 10 and CIFAR 100-20.
First, we compare the impact of different possible pairwise labeling methods in the feature space.
Second, as one of our key contribution is to choose the space where the pairwise labeling is performed, we test doing so at the level of features and predictions (\emph{i.e.} after the linear classifier but before the softmax layer like DEC~\cite{xie2016unsupervised} or DAC~\cite{chang2017deep}).
Third, we analyse the importance of data augmentation in clustering raw images.
Results are reported in \cref{tab:ablation} and discussed next.

\paragraph{Pairwise similarity.}

We compare, in feature space, pairwise labeling methods based on $L_2$ distance, cosine similarity, kNN and symmetric SNE as described in \cref{table:pairwise_lab}. For kNN, we set the number of neighbors $k$ to 20 and 10 for CIFAR 10 and CIFAR 100-20 respectively.
For the cosine similarity, we use respectively thresholds 0.9 and 0.95.
For the $L_2$ distance, we ran a grid search between 0 and 2 to find an optimal threshold.
For SNE, we set the threshold to 0.01 and the temperature to 1 and 0.5, for CIFAR 10 and CIFAR 100-20 respectively. Further details about the hyperparameters are available in the supplementary.
We observe that kNN, SNE and cosine similarity perform very well on CIFAR 10 with values around $81\%$. It is interesting to note that cosine similarity performs noticeably worse than kNN and SNE on CIFAR 100-20 with around 6 points less.
We also notice that $L_2$ distance performs consistently worse than the other labeling methods.
We can conclude that kNN and SNE are the best labeling methods empirically with consistent performance on these two datasets.

\paragraph{Feature space embedding.}

Instead of using these labeling methods before the linear classifier, we apply them after it.
In this case, our overall approach becomes more similar to standard pseudo-labeling methods such as~\cite{chang2017deep, lee2013pseudo, xie2016unsupervised}, which aim to match the network predictions output with a `sharper' version of it.
We observe that the performance drops considerably for all labeling methods with an average decrease of 16.3 points for CIFAR 10 and 10.6 points for CIFAR 100-20.
Hence, this shows empirically that where pseudo labeling is applied plays a major role in clustering effectiveness and that labeling at the feature space level is noticeably better than doing so at the prediction space level.

\paragraph{Data augmentation.}

We compare RICAP, MixUp, and the case without data composition (denoted as None).
As can it can be seen in table~\ref{tab:ablation}, data composition is crucial for CIFAR 10 where RICAP and MixUp surpass None by respectively $28$ and $22$ points.
On CIFAR 100-20, the differences are smaller but using data composition still brings a clear improvement with a $5.1$ points increase when using RICAP.
Interestingly, RICAP clearly outperforms MixUp in both cases. 

\begin{table}
  \caption{{\bf Ablation study.} We analyse the effect of different pairwise labeling methods but also the impact of where the labeling is done (feature vs prediction space). We also show the paramount importance of data augmentation for clustering some datasets like CIFAR 10.}
  \label{tab:ablation}
  \centering
  \footnotesize
  \begin{tabular}{lcccccccccc}
    \toprule
    & \multicolumn{4}{c}{Pairwise labeling} & \multicolumn{3}{c}{Using the pred. space}  & \multicolumn{3}{c}{Data augmentation}                   \\
    \cmidrule(lr){2-5} \cmidrule(lr){6-8} \cmidrule(lr){9-11}
    & $L_2$ & Cosine     & kNN     & SNE  & Cosine & kNN & SNE & RICAP & MixUp & None \\
    \midrule
    CIFAR 10 & 70.2 & 81.1  & \textbf{81.7} & 81.5 & 63.7 & 64.7 & \textbf{67.0} & \textbf{81.7} & 75.3 & 53.7    \\
    CIFAR 100-20 & 26.1 & 34.4 & \textbf{42.3} & 40.4 & 20.4 & \textbf{32.8} & 30.4 & \textbf{42.3} & 37.1 & 35.4   \\
    \bottomrule
  \end{tabular}
\end{table}

\FloatBarrier
\section{Conclusions}
\label{sec:conclusion}
We have proposed a novel deep clustering method, LSD-C.
Our method establishes pairwise connections at the feature space level among different data points in a mini-batch. These on-the-fly pairwise connections are then used as targets by our loss to regroup samples into clusters.
In this way, our method can effectively learn feature representation together with the cluster assignment.
In addition, we also combine recent self-supervised representation learning with our clustering approach to bootstrap the representation before clustering begins.
Finally, we adapt data composition techniques to the pairwise connections setting, resulting in a very large performance boost
 Our method substantially outperforms existing approaches in various public benchmarks, including CIFAR 10/100-20, STL 10, MNIST and Reuters 10K.

\section*{Broader Impact}
Our method considers the task of unsupervised clustering from unlabeled data. We mainly consider two types of data: images and text document. While we make significant advances in terms of clustering accuracy compared to previous work, we believe the data we used to be at low risk since we consider datasets wide-spread around the community for sometimes decades.

While the data we used are not at risk we believe there is an inherent risk of misuse with clustering particularly when learnt from raw data. As any learning algorithm the clustering also depends on the data bias and could lead to misinformation or misinterpretation of results obtained from our model. 

However we believe our method and clustering in general to be of interest for future years as it would reduce the need of heavy data annotations and processing.

\section{Acknowledgments}
We thank Kevin Scaman for his very useful comments. This work is supported by the EPSRC Programme Grant Seebibyte  EP/M013774/1, Mathworks/DTA DFR02620, and ERC IDIU-638009.

{\small\bibliographystyle{plain}
\bibliography{ms}}
\FloatBarrier
\newpage
\appendix{\centering
{\LARGE\bf LSD-C\@: Linearly Separable Deep Clusters\\ Supplementary Material \par}}

\vspace*{30mm}

In this supplementary material, we provide our implementation details, the confusion matrices on CIFAR 10 using our method with kNN labeling and some additional ablation studies. We also include the code to run our method on CIFAR 10 together with the network pretrained with RotNet~\cite{gidaris2018unsupervised}.

\section{Implementation details}

\textbf{Self-supervised pretraining.} We train the RotNet~\cite{gidaris2018unsupervised} (i.e.\ predicting the rotation applied to the image among four possibilities: 0$^\circ$, 90$^\circ$, 180$^\circ$, and 270$^\circ$) on all datasets with the same configuration.  Following the authors' released code, we train for 200 epochs using a step-wise learning rate starting at 0.1 which is then divided by 5 at epochs 60, 120, and 160. 

\textbf{Main LSD-C models.}
After the self-supervised pretraining step, following~\cite{han2020automatically} we freeze the first three macro-blocks of the ResNet-18~\cite{he2016deep} as the RotNet training provides robust early filters. We then train the last macro-block and the linear classifier using our clustering method. For all the experiments, we use a batch size of 256. We summarize in~\cref{tab:hyperparam} all the hyperparameters for the different datasets and labeling methods.

\begin{table}[!htbp]
  \caption{{\bf Hyperparameters.} Optimizer, ramp-up function and parameters of different labeling methods on different datasets.}
  \label{tab:hyperparam}
  \centering
  \footnotesize
  \begin{tabular}{lcccccccccc}
    \toprule
    & \multicolumn{4}{c}{Optimizer} &  \multicolumn{2}{c}{Ramp-up} & \multicolumn{1}{c}{Cosine}  & \multicolumn{2}{c}{SNE} & \multicolumn{1}{c}{kNN}                   \\
    \cmidrule(lr){2-5} \cmidrule(lr){6-7} \cmidrule(lr){8-8} \cmidrule(lr){9-10} \cmidrule(lr){11-11} 
    & Type & Epochs & LR steps & LR init & $\lambda$ & T & $\tau$ & $\tau$ & Temp & k  \\
    \midrule
    CIFAR 10  & SGD & 220 & [140, 180]  & 0.1 & 5 & 100 & 0.9 & 0.01 & 1.0 & 20     \\
    CIFAR 100-20 & SGD & 200 & 170  & 0.1 & 25 & 150 & 0.95 & 0.01 & 0.5 & 10    \\
    STL 10 & SGD & 200 & [140, 180] & 0.1 & 5 & 50 & - & 0.01 & 0.5 & -  \\
    MNIST & SGD & 15 & - & 0.1 & 5 & 50 & - & - & - & 10 \\
    Reuters 10K & Adam & 75 & - & 0.001 & 25 & 100 & - & - & - & 5   \\
    \bottomrule
  \end{tabular}
\end{table}

\textbf{Data augmentation techniques.} We showed in the main paper that data composition techniques like RICAP~\cite{takahashi2018ricap} and MixUp~\cite{zhang2017mixup} are highly beneficial to our method. For RICAP, we follow the authors' instructions to sample the width and height of crops for each minibatch permutation by using a Beta(0.3, 0.3) distribution. Regarding MixUp, we note that using a Beta(0.3, 0.3) distribution for the mixing weight works better in our case than the Beta(1.0, 1.0) advised for CIFAR 10 in the MixUp paper. Furthermore, we have to decrease the weight decay to $10^{-4}$ to make MixUp work.

\textbf{Miscellaneous.} Our method is implemented with PyTorch 1.2.0~\cite{pytorch2019}. Our experiments were run on NVIDIA Tesla M40 GPUs and can run on a single GPU with 12 GB of RAM.

\FloatBarrier
\section{Confusion matrices on CIFAR 10}

\begin{figure}[!htbp]
\centering
\begin{subfigure}{.49\textwidth}
  \centering
  \includegraphics[width=0.8\linewidth]{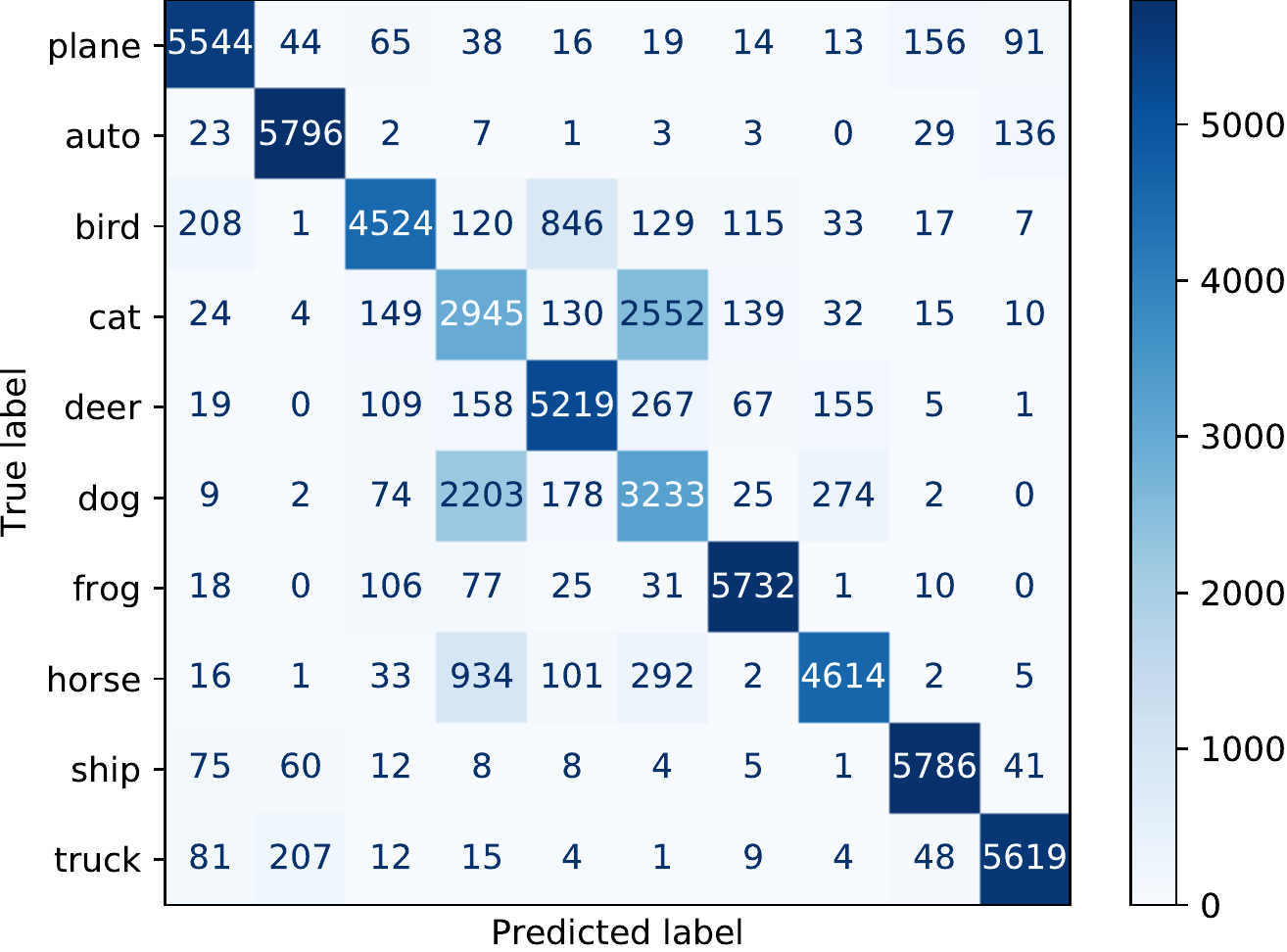}
  \caption{All samples}
  \label{fig:all_confusion}
\end{subfigure}%
\begin{subfigure}{.49\textwidth}
  \centering
  \includegraphics[width=0.8\linewidth]{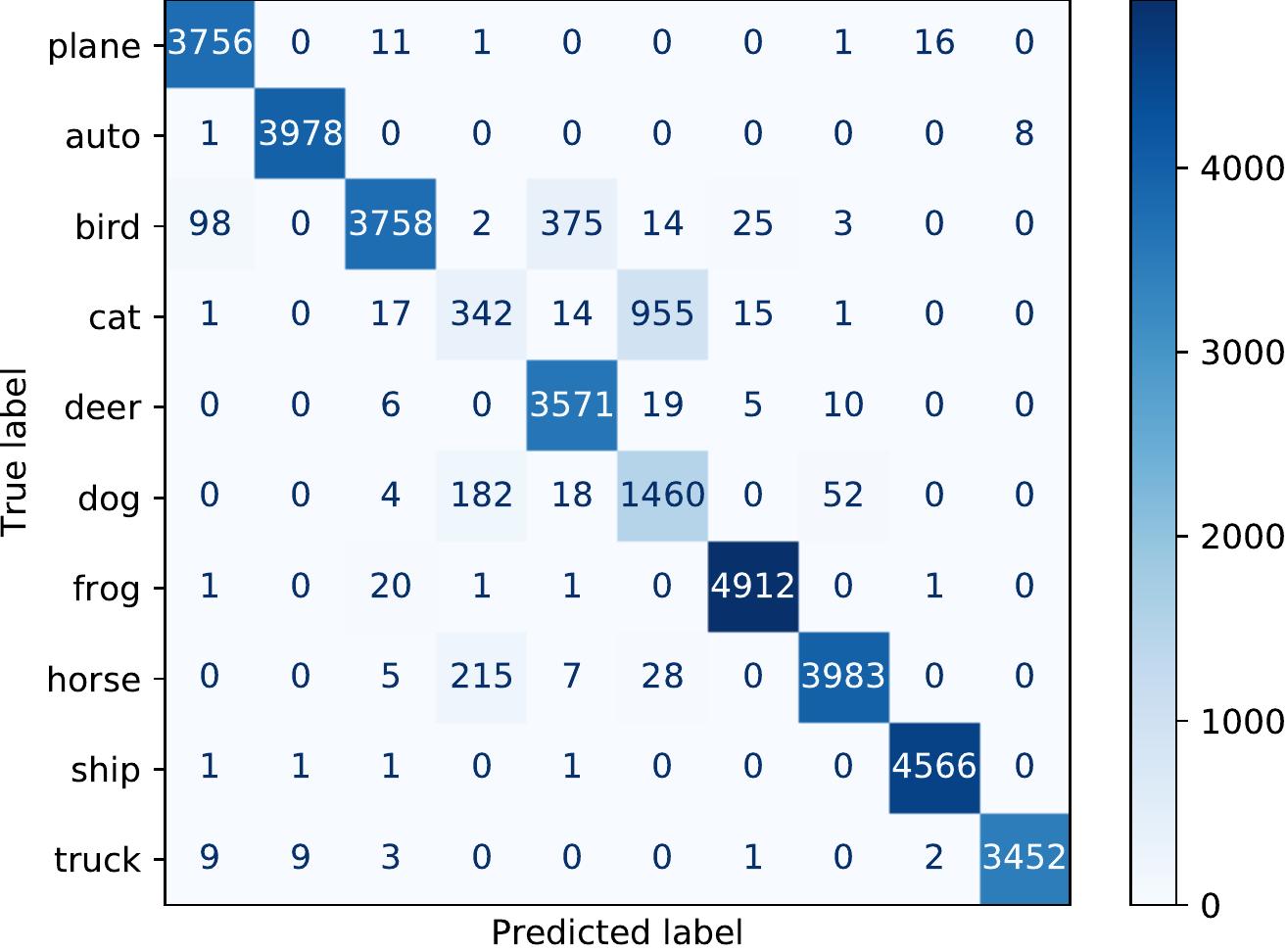}
  \caption{Confident samples}
  \label{fig:confident_confusion}
\end{subfigure}
\caption{\textbf{Confusion matrices on CIFAR 10 using our method with kNN labeling.} Figure~\ref{fig:all_confusion} shows that most of the errors are due to the "cat" and "dog" classes. When taking the samples with prediction above 0.9 (60\% of the samples) in Figure~\ref{fig:confident_confusion}, there are less than 2000 predictions on classes "cat" and "dog" whereas there are more than 3500 for each of the other classes. Our method manages to ignore the problematic classes when taking the confident samples. Indeed, the accuracy for confident samples is 94.0\%.}
\label{fig:confusion}
\end{figure}

In~\cref{fig:confusion}, we show some confusion matrices on CIFAR 10 to analyse how our clustering method performs on the different classes. We notice that there are 8 confident clusters with a very high clustering accuracy of $94.0\%$ for confident samples. The "dog" and "cat" clusters are not well identified possibly due to a huge intra-class variation of the samples. 

\FloatBarrier
\section{Additional ablation studies}

We report in~\cref{tab:add_ablation} the results of some additional ablation studies to evaluate the impact of more components of our method. For example, we apply K-means~\cite{macqueen1967some} on the feature space of the pretrained RotNet model and we note very poor performance on CIFAR 10 and CIFAR 100-20. We can conclude that before training with our clustering loss, the desired clusters are not yet separated in the feature space. After training with our clustering loss, the clusters can be successfully separated. Moreover, if we only use the clustering loss and drop the consistency MSE loss, the performance decreases on both CIFAR 10 and CIFAR 100-20 by  1.5 and 1.3 points respectively, showing that the MSE provides a moderate but clear gain to our method. Finally, if we replace the linear classifier by a 2-layer classifier (i.e.\ this corresponds to a non-linear separation of clusters in the feature space), it results in a small improvement on CIFAR 10 but a clear decrease of 1.9 points on CIFAR 100-20. Hence using a linear classifier provides more consistent results across datasets.

\begin{table}[!htbp]
  \caption{{\bf Additional ablation studies.}  From the first column, we observe that the desired clusters are not yet separated in the feature space after the RotNet pretraining. The second column shows that the MSE consistency loss provides a boost of more than 1 point to our method. Finally, we see that using a non-linear classifier harms the performance on CIFAR 100-20.}
  \label{tab:add_ablation}
  \centering
  \footnotesize
  \begin{tabular}{lcccccccccc}
    \toprule
    & K-means + RotNet & Ours (kNN) & Ours (kNN) w/o MSE & Ours (kNN) w/ non-lin.  \\
    \midrule
    CIFAR 10  & 14.3 & 81.7 & 80.2 & 82.0      \\
    CIFAR 100-20 & 9.1 & 40.5 & 39.2 & 38.6    \\
    \bottomrule
  \end{tabular}
\end{table}

\end{document}